%% file: main.tex
\documentclass{article}
\usepackage{spconf,amsmath,graphicx}

\title{Multi-Sample Online Learning for Spiking Neural Networks based on Generalized Expectation Maximization}
\twoauthors
{Hyeryung Jang\sthanks{This work was done when H. Jang was with King's College London.}}
{Dept. of Artificial Intelligence, \\ Dongguk University, South Korea}
{Osvaldo Simeone\sthanks{This work was supported by the European Research Council (ERC) under the European Union's Horizon 2020 research and innovation programme (grant agreement No. 725731).}}
{
CTR, Dept. of Engineering, \\ King’s College London, United Kingdom.}

\usepackage{amsmath}
\usepackage{amsfonts}
\usepackage{amssymb}
\usepackage{amsthm}
\usepackage{color}
\usepackage{array}
\usepackage{cite}
\usepackage{enumerate}
\usepackage{graphicx}
\usepackage[hang]{subfigure}
\usepackage{epstopdf}
\usepackage{epsfig}
\usepackage{footmisc}
\usepackage{amsbsy}
\usepackage{bm}
\usepackage{cases}
\usepackage{multirow}
\usepackage{algorithmic}
\usepackage{comment}
\usepackage{times}
\usepackage{paralist}
\usepackage[ruled]{algorithm2e}
\usepackage{setspace}

\input{mymath}

\begin{document}
\ninept
\maketitle
%

%
\begin{abstract}
Spiking Neural Networks (SNNs) offer a novel computational paradigm that captures some of the efficiency of biological brains by processing through binary neural dynamic activations. 
Probabilistic SNN models are typically trained to maximize the likelihood of the desired outputs by using unbiased estimates of the log-likelihood gradients. 
While prior work used single-sample estimators obtained from a single run of the network, this paper proposes to leverage multiple compartments that sample independent spiking signals while sharing synaptic weights. 
The key idea is to use these signals to obtain more accurate statistical estimates of the log-likelihood training criterion, as well as of its gradient. 
The approach is based on generalized expectation-maximization (GEM), which optimizes a tighter approximation of the log-likelihood using importance sampling. 
The derived online learning algorithm implements a three-factor rule with global per-compartment learning signals. 
Experimental results on a classification task on the neuromorphic MNIST-DVS data set demonstrate significant improvements in terms of log-likelihood, accuracy, and calibration when increasing the number of compartments used for training and inference. 
\end{abstract}

\vspace{-0.1cm}
\input{introduction}

\vspace{-0.1cm}
\input{model}

\input{gem_learning}
\vspace{-0.1cm}
\input{experiments}

\vspace{-0.1cm}
\input{conclusion}



\newpage
\bibliographystyle{IEEEbib}
\bibliography{ref}

\end{document}

%% file: mymath.tex
\newcommand{\bmx}{{\bm x}}

\newcommand{\bmh}{{\bm h}}
\newcommand{\bms}{{\bm s}}
\newcommand{\bmz}{{\bm z}}

\newcommand{\bmv}{{\bm v}}

\newcommand{\bmtheta}{\bm{\theta}}

\newcommand{\bmsigma}{\bm{\sigma}}

\newcommand{\set}[1]{\ensuremath{\mathcal #1}}
\newcommand{\grad}[1]{\nabla #1}

\newcommand{\textX}{{\text{X}}}
\newcommand{\textH}{{\text{H}}}

\newcommand{\overra}{\overrightarrow}
\newcommand{\overla}{\overleftarrow}
\newcommand{\CNtoCP}{\mathsf{C}_{\text{N} \rightarrow \text{CP}}}
\newcommand{\CCPtoN}{\mathsf{C}_{\text{CP} \rightarrow \text{N}}}

\newcommand\GEMVLSNN{{GEM-VLSNN}}

%% file: introduction.tex
\section{Introduction}
\label{sec:intro}

\vspace{-0.1cm}
Many of the recent breakthroughs towards solving complex tasks have relied on learning and inference algorithms for Artificial Neural Networks (ANNs) that have prohibitive energy and time consumption for implementation on battery-powered devices \cite{hao2019training, strubell2019energy}. This has motivated a renewed interest in exploring the use of Spiking Neural Networks (SNNs) to capture some of the efficiency of biological brains for information encoding and processing \cite{mead1990neuromorphic}. Experimental evidence has confirmed the potential of SNNs in yielding energy consumption savings as compared to ANNs in many tasks of interest, such as keyword spotting \cite{blouw2019benchmarking}, while also demonstrating computational advantages in dynamic environments \cite{sebastian2020memory}.

Inspired by biological brains, SNNs consist of neural units that process and communicate over recurrent computing graphs via sparse binary spiking signals over time, rather than via real numbers \cite{neftci2019surrogate}. Spiking neurons store and update a state variable, the membrane potential, that evolves over time as a function of past spike signals from pre-synaptic neurons. Most implementations are based on deterministic spiking neurons that spike when the membrane potential crosses a threshold. These models can be trained via various approximations of back-propagation through time (BPTT). The approximations aim at dealing with the non-differentiability of the threshold function, and at reducing the complexity of BPTT, making it possible to implement local update rules that do not require global backpropagation paths through the computation graph.

As an alternative, probabilistic spiking neural models based on the generalized linear model (GLM) for spiking neurons can be trained by directly maximizing a lower bound on the likelihood of the SNN producing desired outputs \cite{jimenez2014stochastic, brea2013matching, jang19:spm, zenke2018superspike}. This maximization over the synaptic weights typically uses an unbiased estimate of the gradient of the log-likelihood that leverages a single sample from the output of the neurons. The resulting learning algorithm implements an online learning rule that is local apart from a global learning signal.

\begin{figure}[t!]
    \centering
    \includegraphics[height=0.26\columnwidth]{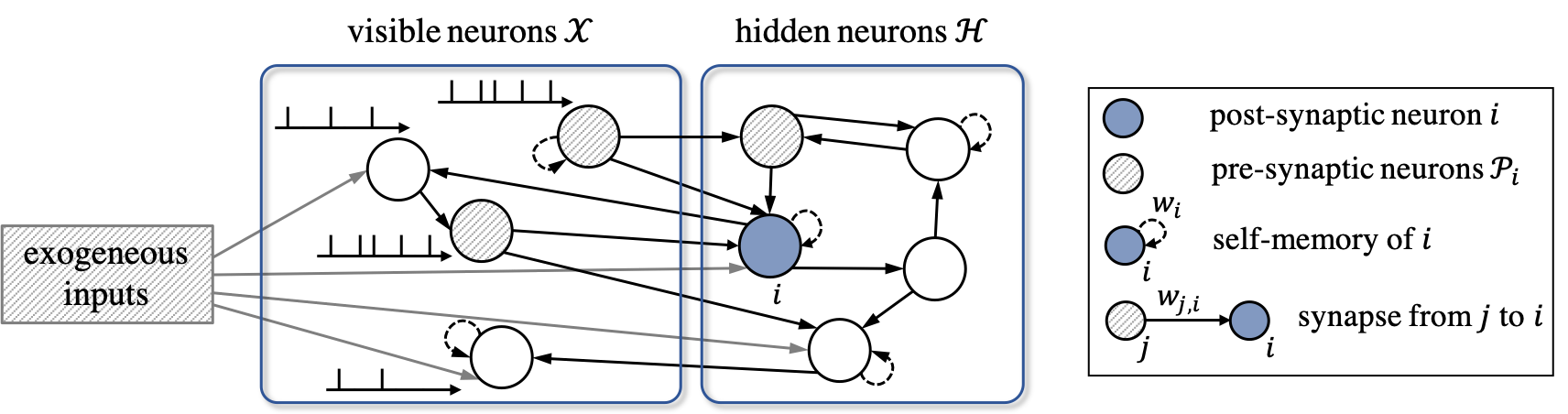}
    \vspace{-0.2cm}
    \caption{Architecture of an SNN with exogeneous inputs and $|\set{X}| = 4$ visible and $|\set{H}| = 5$ hidden spiking neurons -- the directed links between two neurons represent synaptic dependencies, while the self-loop links represent self-memory. Neurons and synapses run $K$ compartments with independent random number generators and shared weights. 
    }
    \label{fig:model-topology}
    \vspace{-0.55cm}
\end{figure}

In this paper, as illustrated in Fig.~\ref{fig:model-topology}, 
we explore a more general SNN model in which each spiking neuron has multiple compartments. Each compartment tracks a distinct membrane potential, with all compartments sharing the same synaptic weights. This architecture would provide no benefits with deterministic models, since all compartments would produce the same outputs. This is not the case under probabilistic neural models when the compartments use independent random number generators to generate spikes. The independent spiking outputs across the compartments can be leveraged in two ways: {\em (i)} during inference, the multiple outputs can be used to robustify the decision and to quantify uncertainty; and {\em (ii)} during training, these signals evaluate a more accurate estimate of the log-likelihood learning criterion and of the corresponding gradient. In this paper, we propose a multi-sample online learning rule that leverages a multi-compartment probabilistic SNN model. The proposed approach adopts generalized expectation-maximization (GEM) \cite{bishop2006pattern, simeone2018brief, neal1998view, tang2013sfnn}, and it uses multiple samples drawn independently from the compartments to approximate the log-likelihood bound via importance sampling. This yields a better statistical estimate of the log-likelihood and of its gradient as compared to conventional single-sample estimators. Experimental results on a neuromorphic data set demonstrate the advantage of leveraging multiple compartments in terms of accuracy and calibration \cite{guo2017calibration}. 


%% file: model.tex
\section{Multi-Compartment SNN Model}
\label{sec:model}


\vspace{-0.1cm}
A $K$-compartment SNN model is defined as a network connecting a set $\set{V}$ of spiking neurons via an arbitrary directed graph, which may have cycles, with each neuron and synapse having $K$ compartments. As illustrated in Fig.~\ref{fig:model-topology}, each neuron $i \in \set{V}$ receives the signals emitted by the set $\set{P}_i$ of pre-synaptic neurons connected to it through directed links, known as synapses. Each $k$th synaptic compartment for a synapse $(j,i)$ processes the output of the $k$th compartment of pre-synaptic neuron $j$; and its output is in turn processed by the $k$th compartment of the post-synaptic neuron $i$, for $k=1,2,\ldots,K$. 

Following a discrete-time implementation of the probabilistic generalized linear neural model (GLM) for spiking neurons \cite{pillow2008spatio, jang19:spm}, at any time $t=1,2,\ldots$, the $k$th compartment of spiking neuron $i$ outputs a binary value $s_{i,t}^k \in \{0,1\}$, with ``$1$'' representing the emission of a spike, for $k=1,\ldots,K$. We collect in vector $\bms_t^k = (s_{i,t}^k: i \in \set{V})$ the spikes of the $k$th compartment emitted by all neurons $\set{V}$ at time $t$, and denote by $\bms_{\leq t}^k = (\bms_1^k, \ldots, \bms_t^k)$ the spike sequences of all neurons processed by the compartment up to time $t$. Each post-synaptic neuron $i$ receives past input spike signals $\bms_{\set{P}_i, \leq t-1}^k$ from the set $\set{P}_i$ of pre-synaptic neurons through the $k$th compartment of synapses. With some abuse of notation, we include exogeneous inputs to a neuron $i$ in the set $\set{P}_i$ of pre-synaptic neurons, see Fig.~\ref{fig:model-topology}. 

For the $k$th compartment, independently of the other compartments, the instantaneous spiking probability of neuron $i$ at time $t$, conditioned on the value of the {\em membrane potential}, is defined as 
\begin{align} \label{eq:spike-prob-ind}
    p_{\bmtheta_i}(s_{i,t}^k=1| u_{i,t}^k) = \sigma(u_{i,t}^k),
\end{align}
where $\sigma(x) = (1+\exp(-x))^{-1}$ is the sigmoid function and the membrane potential $u_{i,t}^k$ summarizes the effect of the past spike signals $\bms_{\set{P}_i, \leq t-1}^k$ from pre-synaptic neurons and of its past activity $\bms_{i,\leq t-1}^k$. Note that each compartment stores and updates a distinct membrane potential. From \eqref{eq:spike-prob-ind}, the negative log-probability of the output $s_{i,t}^k$ corresponds to the binary cross-entropy loss
\begin{align} \label{eq:log-prob-ind}
    - &\log p_{\bmtheta_i}(s_{i,t}^k | u_{i,t}^k) = \ell\big( s_{i,t}^k, \sigma(u_{i,t}^k)\big) \cr 
    &\qquad := -s_{i,t}^k \log \sigma(u_{i,t}^k) - (1-s_{i,t}^k) \log (1-\sigma(u_{i,t}^k)).
\end{align}
The joint probability of the spike signals $\bms_{\leq T}^k$ up to time $T$ is defined using the chain rule as $p_{\bmtheta}(\bms_{\leq T}^k) = \prod_{t=1}^T \prod_{i \in \set{V}} p_{\bmtheta_i}(s_{i,t}^k|u_{i,t}^k)$, where $\bmtheta = \{\bmtheta_i\}_{i \in \set{V}}$ is the model parameters, with $\bmtheta_i$ being the local model parameters of neuron $i$ as detailed below.

The membrane potential is obtained as the output of spatio-temporal synaptic filter $a_t$ and somatic filter $b_t$. Specifically, each synapse $(j,i)$ from a pre-synaptic neuron $j \in \set{P}_i$ to a post-synaptic neuron $i$ computes the synaptic trace $\overra{s}_{j,t}^k = a_t \ast s_{j,t}^k$, while the somatic trace of neuron $i$ is computed as $\overla{s}_{i,t}^k = b_t \ast s_{i,t}^k$, where we denote by $f_t \ast g_t = \sum_{\delta} f_{\delta} g_{t-\delta}$ the convolution operator. The membrane potential $u_{i,t}^k$ of neuron $i$ at time $t$ is finally given as the weighted sum 
\begin{align} \label{eq:membrane-potential-k}
    u_{i,t}^k = \sum_{j \in \set{P}_i} w_{j,i} \overra{s}_{j,t-1}^k + w_i \overla{s}_{i,t-1}^k + \vartheta_i, 
\end{align}
where $w_{j,i}$ is a synaptic weight of the synapse $(j,i)$; $w_i$ is the self-memory weight; and $\vartheta_i$ is a bias, with the local model parameters $\bmtheta_i = \{\{w_{j,i}\}_{j \in \set{P}_i}, w_i, \vartheta_i\}$ being shared among all compartments of neuron $i$.

%% file: gem_learning.tex
\section{Training}
\label{sec:learning}

\vspace{-0.1cm}
During training, the shared model parameters $\bmtheta$ are adapted jointly across all compartments with the goal of maximizing the log-likelihood that a subset $\set{X} \subseteq \set{V}$ of ``visible'' neurons outputs desired spiking signals. The desired samples are specified by the training data as spiking sequences $\bmx_{\leq T}$ for some $T > 0$, in response to given exogeneous inputs. Mathematically, the maximum log-likelihood (ML) problem can be written as 
\begin{align} \label{eq:ml}
    \min_{\bmtheta}~ - \log p_{\bmtheta}(\bmx_{\leq T}),
\end{align}
where $p_{\bmtheta}(\bmx_{\leq T})$ is the probability of the desired output for any of the compartments. To evaluate the log-likelihood $\log p_{\bmtheta}(\bmx_{\leq T}) = \log \sum_{\bmh_{\leq T}} p_{\bmtheta}(\bmx_{\leq T}, \bmh_{\leq T})$, one needs to average over the spiking signals $\bmh_{\leq T}$ of ``hidden'', or latent, neurons in the complement set $\set{H} = \set{V} \setminus \set{X}$.

In this section, we propose an online learning rule that leverages the available $K$ compartments by following the generalized expectation-maximization (GEM) introduced in \cite{tang2013sfnn}. Throughout the paper, we define the temporal average operator of a time sequence $\{f_t\}_{t \geq 1}$ with some constant $\kappa \in (0,1)$ as $\langle f_t \rangle_{\kappa} = \kappa \cdot \langle f_{t-1} \rangle_{\kappa} + f_t$ with $\langle f_0 \rangle_{\kappa} = 0$.

{\bf GEM-VLSNN: Generalized EM Variational online Learning for SNNs.} 
Following the principles of GEM \cite{bishop2006pattern, simeone2018brief, neal1998view, tang2013sfnn}, given the current iterate $\bmtheta_{\text{old}}$, we tackle the problem of minimizing at each time $t$ the following approximate upper bound on the objective in \eqref{eq:ml} \cite{bishop2006pattern, jang19:spm}
\begin{align} \label{eq:gem-objective}
    L_{\bmx_{\leq t}}(\bmtheta, \bmtheta_\text{old}) = \mathbb{E}_{p_{\bmtheta_\text{old}}(\bmh_{\leq t} | \bmx_{\leq t})}\bigg[ \sum_{t'=0}^{t-1} \gamma^{t'} \sum_{i \in \set{V}} \ell\big( s_{i,t-t'}, \sigma(u_{i,t-t'})\big) \bigg],
\end{align}
where $\gamma \in (0,1)$ is a discount factor and $p_{\bmtheta_{\text{old}}}(\bmh_{\leq t}|\bmx_{\leq t})$ is the posterior distribution. The bound is exact for $\gamma = 1$, and we have added the discounted average in \eqref{eq:gem-objective} in order to obtain online rules. The posterior distribution $p_{\bmtheta_{\text{old}}}(\bmh_{\leq t}|\bmx_{\leq t})$ is generally intractable, and, following \cite{jimenez2014stochastic, brea2013matching, jang19:spm}, we approximate it with the ``causally conditioned'' distribution $p_{\bmtheta_{\text{old}}^\textH}(\bmh_{\leq t}||\bmx_{\leq t-1})$, where we have used the notation \cite{kramer1998directed}
\begin{align} \label{eq:causally-cond}
    p_{\bmtheta^\textH}(\bmh_{\leq t}||\bmx_{\leq t-1}) 
    = \prod_{t'=1}^t \prod_{i \in \set{H}} p_{\bmtheta_i}(h_{i,t'}|u_{i,t'}),
\end{align}
and we have denoted by $\bmtheta^\textX = \{\bmtheta_i\}_{i \in \set{X}}$ and $\bmtheta^\textH = \{\bmtheta_i\}_{i \in \set{H}}$ the collection of model parameters for visible and hidden neurons respectively. Then, following GEM, we use $K$ independent samples $\bmh_{\leq t}^{1:K} = \{\bmh_{\leq t}^k\}_{k=1}^K$ drawn from the distribution $p_{\bmtheta_{\text{old}}^\textH}(\bmh_{\leq t}^{1:K} || \bmx_{\leq t-1}) = \prod_{k=1}^K p_{\bmtheta_{\text{old}}^\textH}(\bmh_{\leq t}^k || \bmx_{\leq t-1})$ to carry out the marginalization via importance sampling. Accordingly, we approximate the loss function \eqref{eq:gem-objective} as
\begin{align} \label{eq:gem-elbo}
    &L_{\bmx_{\leq t}}(\bmtheta, \bmtheta_\text{old}) \cr
    &\quad \approx \frac{1}{K} \sum_{k=1}^K \frac{p_{\bmtheta_{\text{old}}}(\bmh_{\leq t}^k|\bmx_{\leq t})}{p_{\bmtheta_{\text{old}}^\textH}(\bmh_{\leq t}^k||\bmx_{\leq t-1})} \sum_{t'=0}^{t-1} \gamma^{t'} \sum_{i \in \set{V}} \ell\big( s_{i,t-t'}^k, \sigma(u_{i,t-t'}^k) \big) \cr
    &\quad \approx \sum_{k=1}^K \bmsigma_{\text{SM}}^k \Big( \bmv_{\bmtheta_{\text{old}}^\textX,t} \Big) \cdot \sum_{t'=0}^{t-1} \gamma^{t'} \sum_{i \in \set{V}} \ell\big( s_{i,t-t'}^k, \sigma(u_{i,t-t'}^k)\big) \cr
    &\quad := L_{\bmx_{\leq t}}^K(\bmtheta, \bmtheta_\text{old}).
\end{align}
The approximate upper bound $L_{\bmx_{\leq t}}^K(\bmtheta, \bmtheta_{\text{old}})$ in \eqref{eq:gem-elbo} is obtained by approximating the {\em importance weights} of the samples as
\begin{align} \label{eq:iw-approx}
    &\frac{1}{K} \cdot \frac{ p_{\bmtheta_{\text{old}}}(\bmh_{\leq t}^k | \bmx_{\leq t}) }{ p_{\bmtheta_{\text{old}}^\text{H}}(\bmh_{\leq t}^k || \bmx_{\leq t-1})} 
    = \frac{p_{\bmtheta_{\text{old}}}(\bmx_{\leq t}, \bmh_{\leq t}^k)}{K \cdot  p_{\bmtheta_{\text{old}}}(\bmx_{\leq t}) \cdot p_{\bmtheta_{\text{old}}^\text{H}}(\bmh_{\leq t}^k || \bmx_{\leq t-1})} \cr 
    &\qquad \stackrel{(a)}{\approx} \frac{ p_{\bmtheta_{\text{old}}^\text{X}} (\bmx_{\leq t} || \bmh_{\leq t-1}^k)}{\sum_{k'=1}^K p_{\bmtheta_{\text{old}}^\text{X}}(\bmx_{\leq t} || \bmh_{\leq t-1}^{k'}) } 
    \stackrel{(b)}{\approx} \bmsigma_{\text{SM}}^k\Big( \bmv_{\bmtheta_{\text{old}}^\textX,t} \Big),
\end{align}
where in (a) we have used a Monte Carlo (MC) estimate with $K$ samples $\bmh_{\leq t}^{1:K}$ for $p_{\bmtheta_{\text{old}}}(\bmx_{\leq t}) \approx \frac{1}{K} \sum_{k'=1}^K p_{\bmtheta_{\text{old}}^\textX}(\bmx_{\leq t}||\bmh_{\leq t-1}^{k'})$; and in (b) we have defined $\bmv_{\bmtheta_{\text{old}}^\textX, t} = (v_{\bmtheta_{\text{old}}^\textX,t}^1, \ldots, v_{\bmtheta_{\text{old}}^\textX,t}^K)$ as the vector of log-probabilities of the samples at time $t$ by using temporal averaging operator as
\begin{align} \label{eq:log-iw}
    v_{\bmtheta_{\text{old}}^\textX,t}^k = \Big\langle \log p_{\bmtheta_{\text{old}}^\textX}(\bmx_{\leq t} || \bmh_{\leq t-1}^k) \Big\rangle_{\kappa}
\end{align}
with some constant $\kappa \in (0,1)$, and have defined the SoftMax function $\bmsigma_{\text{SM}}(\cdot)$ as
\begin{align} \label{eq:softmax-iw}
    \bmsigma_{\text{SM}}^k\Big( \bmv_{\bmtheta_{\text{old}}^\textX, t} \Big) = \frac{\exp\big( v_{\bmtheta_{\text{old}}^\textX, t}^k \big)}{\sum_{k'=1}^K \exp\big( v_{\bmtheta_{\text{old}}^\textX, t}^{k'} \big)}, ~\text{for}~ k=1,\ldots,K.
\end{align}
We note that the resulting MC estimate can better capture the inherent uncertainty of the true posterior distribution more precisely with a larger $K$.

\begin{figure}[t!]
    \centering
    \includegraphics[height=0.53\columnwidth]{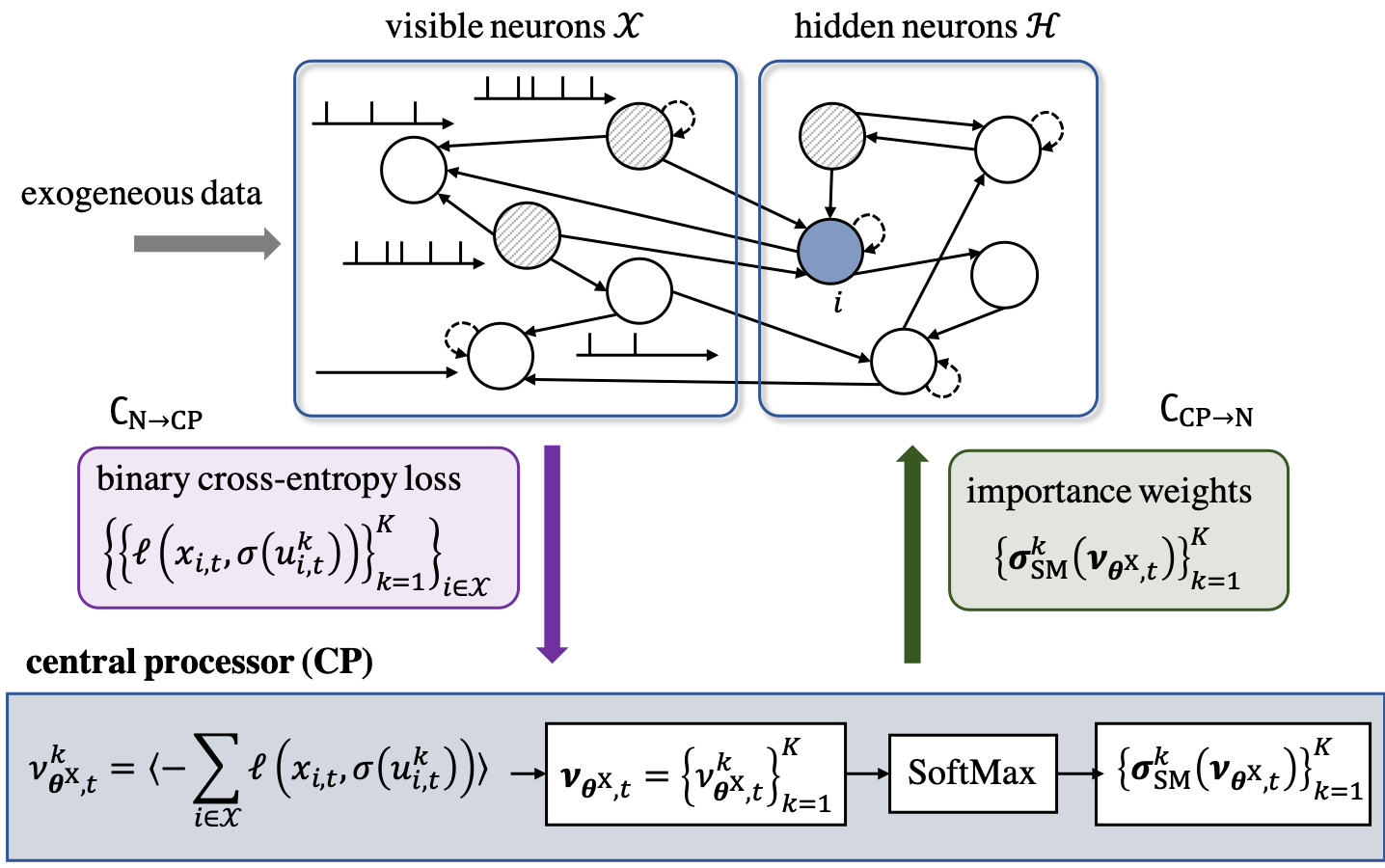}
    \vspace{-0.15cm}
    \caption{Illustration of the training scheme for a $K$-compartment SNN. A central processor (CP) collects information from all compartments of the visible neurons $\set{X}$, with entailing a unicast communication load $\CNtoCP$; computes importance weights of compartments; and sends them to all neurons, with entailing a broadcast communication load $\CCPtoN$, in order to guide the update of model parameters $\bmtheta$.
    }
    \label{fig:model-learning}
    \vspace{-0.5cm}
\end{figure}

An online local update rule, which we refer to as GEM Variational online Learning for SNNs ({\GEMVLSNN}), is obtained by minimizing $L_{\bmx_{\leq t}}^K(\bmtheta, \bmtheta_{\text{old}})$ in \eqref{eq:gem-elbo} via stochastic gradient descent in the direction of the negative gradient $- \grad_{\bmtheta} L_{\bmx_{\leq t}}^K(\bmtheta, \bmtheta_{\text{old}})$. This yields the update rule at time $t$
\begin{align*}
    \bmtheta \leftarrow \bmtheta - \eta \cdot \sum_{k=1}^K \bmsigma_{\text{SM}}^k\big( \bmv_{\bmtheta^\textX,t} \big) \cdot \sum_{t'=0}^{t-1} \gamma^{t'} \sum_{i \in \set{V}} \grad_{\bmtheta} \ell\big( s_{i,t-t'}^k, \sigma(u_{i,t-t'}^k) \big),
\end{align*}
with a learning rate $\eta$. As illustrated in Fig.~\ref{fig:model-learning}, this rule can be implemented as follows. At each time $t=1,2,\ldots$, a central processor (CP) collects the binary cross-entropy values $\{\ell\big( x_{i,t}, \sigma(u_{i,t}^k) \big) \}_{k=1}^K$ of all compartments from all visible neurons $i \in \set{X}$ in order to compute the importance weights of the compartments $\bmv_{\bmtheta^\textX,t}$ as in \eqref{eq:log-iw}, and it computes the normalized weights using the SoftMax function $\bmsigma_{\text{SM}}(\cdot)$ as in \eqref{eq:softmax-iw}. Then, the SoftMax values $\{ \bmsigma_{\text{SM}}^k\big( \bmv_{\bmtheta^\textX,t} \big)\}_{k=1}^K$ are fed back from the CP to all neurons $\set{V}$ (both visible $\set{X}$ and hidden $\set{H}$ neurons). Finally, for each neuron $i \in \set{V}$, {\GEMVLSNN} updates the local model parameters $\bmtheta_i$ as
\begin{align} \label{eq:gem-update}
    w_{j,i} &\leftarrow w_{j,i} - \eta \cdot \sum_{k=1}^K \bmsigma_{\text{SM}}^k\big( \bmv_{\bmtheta^\textX,t} \big) \cdot \Big\langle \big( s_{i,t}^k - \sigma(u_{i,t}^k) \big) \cdot \overra{s}_{j,t-1}^{k} \Big\rangle_\gamma, \cr
    w_i &\leftarrow w_i - \eta \cdot \sum_{k=1}^K \bmsigma_{\text{SM}}^k\big( \bmv_{\bmtheta^\textX,t} \big) \cdot \Big\langle \big( s_{i,t}^k - \sigma(u_{i,t}^k) \big) \cdot \overla{s}_{i,t-1}^k \Big\rangle_\gamma, \cr
    \vartheta_i &\leftarrow \vartheta_i - \eta \cdot \sum_{k=1}^K \bmsigma_{\text{SM}}^k\big( \bmv_{\bmtheta^\textX,t} \big) \cdot \Big\langle s_{i,t}^k - \sigma(u_{i,t}^k) \Big\rangle_\gamma,
\end{align}
where we have used standard expressions for the derivatives of the cross-entropy loss \cite{jang19:spm, jang19:def}, and we set $s_{i,t}^k = x_{i,t}$ for visible neuron $i \in \set{X}$ and $s_{i,t}^k = h_{i,t}^k$ for hidden neuron $i \in \set{H}$.

{\bf Interpreting GEM-VLSNN.} 
The update rule \eqref{eq:gem-update} follows a three-factor format in the sense that it can be written as \cite{fremaux2016neuromodulated}
\begin{align} \label{eq:three-factor}
    w_{j,i} \leftarrow w_{j,i} + \eta \cdot \sum_{k=1}^K \textsf{learning signal}^k \cdot \big\langle \textsf{pre}_j^k \cdot \textsf{post}_i^k \big\rangle,
\end{align}
where $\eta$ is the learning rate. The three-factor rule \eqref{eq:three-factor} sums the contributions from the $K$ compartments, with contribution of the $k$th compartment depending on three factors: $\textsf{pre}_j^k$ and $\textsf{post}_i^k$ respectively representing the activity of the pre-synaptic neuron $j$ and of the post-synaptic neuron $i$ processed by the compartment, and finally $\textsf{learning signal}^k$ determines the sign and magnitude of the contribution of the compartment to the learning update. The update rule \eqref{eq:three-factor} is hence local, with the exception of the learning signal. Specifically, for each neuron $i$, the gradients $\grad_{\bmtheta_i} L_{\bmx_{\leq t}}^K(\cdot)$ contain the post-synaptic error $s_{i,t}^k - \sigma(u_{i,t}^k)$ and post-synaptic somatic trace $\overla{s}_{i,t-1}^k$; the pre-synaptic synaptic trace $\overra{s}_{j,t-1}^k$; and the global learning signals $\{ \bmsigma_{\text{SM}}^k\big( \bmv_{\bmtheta^\textX,t} \big)\}_{k=1}^K$. The importance weights computed using the SoftMax function can be interpreted as the common learning signals for all neurons, with the contribution of each compartment being weighted by $\bmsigma_{\text{SM}}^k(\bmv_{\bmtheta^\textX,t})$. From \eqref{eq:log-iw}-\eqref{eq:softmax-iw}, the importance weight $\bmsigma_{\text{SM}}^k(\bmv_{\bmtheta^\textX,t})$ measures the relative effectiveness of the random realization $\bmh_{\leq t}^k$ of the hidden neurons within the $k$th compartment in reproducing the desired behavior $\bmx_{\leq t}$ of the visible neurons.

{\bf Communication Load.} 
As discussed, {\GEMVLSNN} requires bi-directional communication. As seen in Fig.~\ref{fig:model-learning}, at each time $t$, unicast communication from neurons to CP is required in order to compute the importance weights by collecting information $\{\{\ell\big(x_{i,t}, \sigma(u_{i,t}^k)\big)\}_{k=1}^K\}_{i \in \set{X}}$ from all visible neurons. The resulting unicast communication load is $\CNtoCP = K |\set{X}|$ real numbers. The importance weights $\{\bmsigma_{\text{SM}}^k\big( \bmv_{\bmtheta^\textX,t} \big)\}_{k=1}^K$ are then sent back to all neurons, resulting a broadcast communication load from CP to neurons equal to $\CCPtoN = K(|\set{X}|+|\set{H}|)$ real numbers. As {\GEMVLSNN} requires computation of $K$ importance weights at CP, the communication loads increase linearly to $K$.

%% file: experiments.tex
\section{Experiments} \label{sec:experiments}

\vspace{-0.1cm}
In this section, we evaluate the performance of the proposed scheme {\GEMVLSNN} on classification task defined on the neuromorphic data set MNIST-DVS \cite{serrano2015poker}. For each pixel of an image, binary spiking signals are recorded when the pixel's luminosity changes by more than a given amount, and no event is recorded otherwise. As in \cite{zhao2014feedforward, henderson2015spike}, images are cropped to $26 \times 26$ pixels, and uniform downsampling over time is carried out to obtain $T=80$ time samples per each image. The training and test data set respectively contains $900$ and $100$ examples per each digit, from $0$ to $9$. We focus on a classification task that classifies three digits $\{0,1,2\}$, where the $26 \times 26$ spiking signals encoding an MNIST-DVS image are given as exogeneous inputs. The digit labels are encoded by the neurons in the read-out layer, where the output neuron $c \in \set{X}$ corresponding to the correct label is assigned a desired output spiking signal $x_{c,t} = 1$, while the other neurons $c' \neq c$ are assigned $x_{c',t} = 0$ for $t=1,\ldots,T$. 

We consider a generic, non-optimized network architecture with a set of $|\set{H}| = 200$ fully connected hidden neurons, all receiving the exogeneous inputs as pre-synaptic signals, and a read-out visible layer with $|\set{X}|=3$ neurons, directly receiving pre-synaptic signals from all exogeneous inputs and all hidden neurons without recurrent connections between visible neurons. For synaptic and somatic filters, we choose a set of three raised cosine functions with a synaptic duration of $10$ time steps, following the approach \cite{pillow2008spatio}. We train a $K$-compartment SNN by varying the number $K$, with the learning rate $\eta = 0.001$ and time constants $\kappa=\gamma=0.9$, which have been selected after a non-exhaustive manual search. 

\begin{figure}[t!]
\centering
\includegraphics[height=0.34\columnwidth]{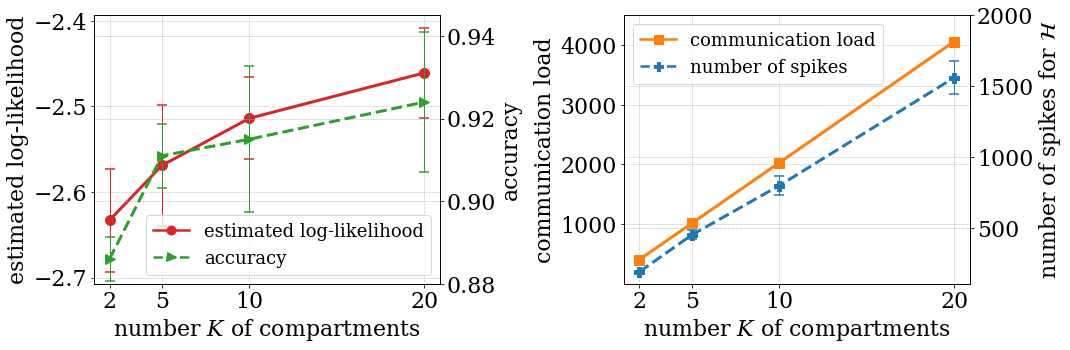} 
\vspace{-0.2cm}
\caption{Classification performance versus $K$, with $95\%$ confidence intervals accounting for the error bars: (Left) estimated log-likelihood for the desired output (solid) and classification accuracy (dashed) on test data set; (Right) broadcast communication load $\CCPtoN$ (solid) and number of spikes emitted by the hidden neurons $\set{H}$ per unit time (dashed) during training.
}
\label{fig:general-classification} 
\vspace{-0.4cm}
\end{figure}

For testing, a $K^I$-compartment SNN with the trained weights is used for inference. First, we measure the marginal log-likelihood $\log p_{\bmtheta}(\bmx_{\leq T})$ for a desired output signal $\bmx_{\leq T}$ via the empirical average over $20$ independent realizations of the hidden neurons. Next, in order to evaluate the classification accuracy, we adopt a standard majority decoding rule: for each compartment $k=1,\ldots,K^I$, the output neuron with the largest number of output spikes for each $\bmx_{\leq T}^k$ is selected, obtaining the decision $\hat{c}^k := \arg\max_{c \in \set{X}} \sum_{t=1}^T x_{c,t}^k$; then the index of the output neuron that receives the most votes is set to the predicted class as $\hat{c} = \arg\max_{c \in \set{X}} z_c$, where $z_c = \sum_{k=1}^{K^I} {\bm 1}_{\{ \hat{c}^k = c \}}$ is the number of votes for class $c$. Finally, we consider {\em calibration} as a performance metric. 
To this end, the prediction probability $\hat{p}$, or confidence, of a decision is derived from the vote count variables $\bmz = (z_c: c \in \set{X})$ using the SoftMax function, i.e., $\hat{p} = \bmsigma_{\text{SM}}^{\hat{c}} \big( \bmz \big)$. The expected calibration error (ECE) measures the difference in expectation between confidence and accuracy, i.e., 
\begin{align} \label{eq:ece}
    \text{ECE} = \mathbb{E}_{\hat{p}} \Big[ \big| \mathbb{P}\big( \hat{c} = c | \hat{p} = p \big) - p \big| \Big].
\end{align}
In \eqref{eq:ece}, the probability $\mathbb{P}\big( \hat{c} = c| \hat{p}=p\big)$ is the probability that $\hat{c}$ is the correct decision for inputs that yield accuracy $\hat{p} = p$. The ECE can be estimated by using quantization and empirical averages as detailed in \cite{guo2017calibration}.

\begin{figure}[t!]
\centering
\includegraphics[height=0.5\columnwidth]{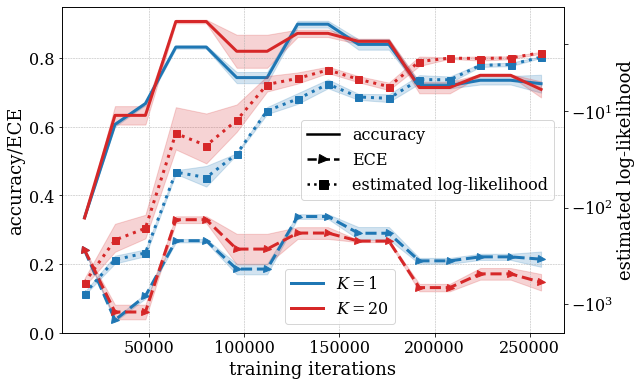} 
\vspace{-0.15cm}
\caption{Estimated log-likelihood, classification accuracy and ECE \eqref{eq:ece} of test data set as a function of processed time samples for different values $K=1,20$ of compartments in training. The accuracy and ECE are measured using $K^I = 2$ compartments. The shaded areas represent $95\%$ confidence intervals.
}
\label{fig:general-ll-ece-gem} 
\vspace{-0.4cm}
\end{figure}

To start, we trained a $K$-compartment SNN and tested the SNN with $K^I = K$ compartments for $K=1,2,5,10,20$. We chose the model with the best performance on the test data set across the iterations, and the corresponding estimated log-likelihood and accuracy are illustrated as a function of $K$ in Fig.~\ref{fig:general-classification}. It can be observed that using more compartments $K$ improves the testing performance due to the optimization of increasingly tighter bound on the training log-likelihood.  Fig.~\ref{fig:general-classification} also shows the broadcast communication load $\CCPtoN$ from CP to neurons increases linearly with $K$, with a larger $K$ implying a proportionally larger number of spikes emitted by the hidden neurons. The proposed {\GEMVLSNN} is seen to enable a flexible trade-off between communication load and energy consumption \cite{merolla2014million}, on the one hand, and testing performance, on the other, by leveraging the availability of $K$ compartments.

We then plot in Fig.~\ref{fig:general-ll-ece-gem} the estimated log-likelihood, accuracy, and ECE \eqref{eq:ece} of test data set as a function of the number of training iterations. An improved test log-likelihood due to a larger $K$ translates into a model that more accurately reproduces conditional probability of outputs given inputs \cite{guo2017calibration, bishop1994mixture}, which in turn enhances calibration. In contrast, accuracy can be improved with a larger $K$ but only if regularization via early stopping is carried out. This points to the fact that the goal of maximizing the likelihood of specific desired output spiking signals is not equivalent to maximizing the classification accuracy.

%% file: conclusion.tex
\section{Conclusion}
\label{sec:conclusion}

\vspace{-0.1cm}
This paper has explored a probabilistic spiking neural model in which spiking neurons run multiple compartments, each generating independent spiking outputs, while sharing the same synaptic weights across compartments. We have proposed a novel multi-sample online learning algorithm for SNNs that leverages multiple compartments to obtain a better statistical estimate of the log-likelihood learning criterion and of its gradient. Experiments on a neuromorphic data set have demonstrated improvements of performance with increasing number of compartments in training and inference. As future work, the proposed learning algorithm can be extended to learning tasks with other reward functions instead of log-likelihood, such as Van Rossum distance \cite{zenke2018superspike, rossum2001novel}, or to networks of spiking Winner-Take-All (WTA) circuits \cite{mostafa2018learning, jang20:vowel}, which process multi-valued spikes.